\documentclass[11pt]{article}

\usepackage[preprint]{acl}
\usepackage{amsmath}
\usepackage[utf8]{inputenc}  
\usepackage{times}
\usepackage{latexsym}
\usepackage{float}
\usepackage{svg}
\usepackage{booktabs}
\usepackage{amssymb}
\usepackage{pifont}
\usepackage{adjustbox}
\usepackage{epstopdf}
\usepackage{hyperref}

\usepackage{multirow}
\usepackage{placeins}

\usepackage[T1]{fontenc}

\usepackage[utf8]{inputenc}
\usepackage{graphicx}
\usepackage{amsmath}

\usepackage{microtype}

\usepackage{inconsolata}

\usepackage{graphicx}

\title{GovRelBench:A Benchmark for Government
Domain Relevance}

\author{Haiquan Wang\textsuperscript{1,2}, \ Yi Chen\textsuperscript{1,2}, \ Shang Zeng\textsuperscript{1,2}, \ Yun Bian\textsuperscript{1,2},Zhe Cui\textsuperscript{\textsuperscript{*\dag} } \\
    \textsuperscript{1}Chengdu Institute of Computer Applications, Chinese Academy of Sciences, Chengdu, China \\
    \textsuperscript{2}University of Chinese Academy of Sciences, Beijing, China \\
    \texttt{\{wanghaiquan22,chenyi24,zengshang19\}@mails.ucas.ac.cn} \\
    \texttt{\{bianyun,cuizhe\textsuperscript{(*)}\}@casit.com.cn} \\
}

\begin{document}
\maketitle
\begin{abstract}
Current evaluations of LLMs in the government domain primarily focus on safety considerations in specific scenarios, while the assessment of the models' own core capabilities, particularly domain relevance, remains insufficient. To address this gap, we propose GovRelBench, a benchmark specifically designed for evaluating the core capabilities of LLMs in the government domain. GovRelBench consists of government domain prompts and a dedicated evaluation tool, GovRelBERT. During the training process of GovRelBERT, we introduce the SoftGovScore method: this method trains a model based on the ModernBERT architecture by converting hard labels to soft scores, enabling it to accurately compute the text's government domain relevance score. This work aims to enhance the capability evaluation framework for large models in the government domain, providing an effective tool for relevant research and practice. Our code and dataset are available at \url{https://github.com/pan-xi/GovRelBench}.
\end{abstract}

\section{Introduction}

Since the advent of ChatGPT~\cite{ouyang2022training} in 2022, the field of Natural Language Processing (NLP) has entered the era of Large Language Models (LLMs). In recent years, research has yielded not only prominent closed-source models like ChatGPT and Claude~\cite{bai2022constitutional} but also a growing number of powerful open-source alternatives such as LLaMA, Qwen, and DeepSeek. The capabilities of these open-source models are rapidly advancing, often approaching the performance levels of their closed-source counterparts, signifying increasingly sophisticated intelligence across the NLP landscape. Concurrently, techniques like instruction tuning~\cite{zhang2023instruction} and parameter-efficient fine-tuning methods such as LoRA~\cite{hu2022lora} have significantly lowered the cost of customizing models. This has successfully facilitated the development of domain-specific models across various sectors, including finance~\cite{lee2025large}, healthcare~\cite{liu2024survey}, and law~\cite{lai2024large}.

When developing models specialized for the government domain, their evaluation is crucial. In addition to general capabilities, evaluating model performance within governmental contexts is particularly critical. However, research on benchmarks for the government domain is currently relatively limited, and existing work~\cite{liu2025evaluation} often prioritizes safety assessments in specific operational scenarios. Furthermore, knowledge in the government domain often lacks universal applicability, as government standards in different countries and regions can vary significantly. Therefore, we propose using the relevance of model output to the government domain as a key metric for evaluating its performance.

\begin{figure*}[t]

    \includegraphics[width=\linewidth]{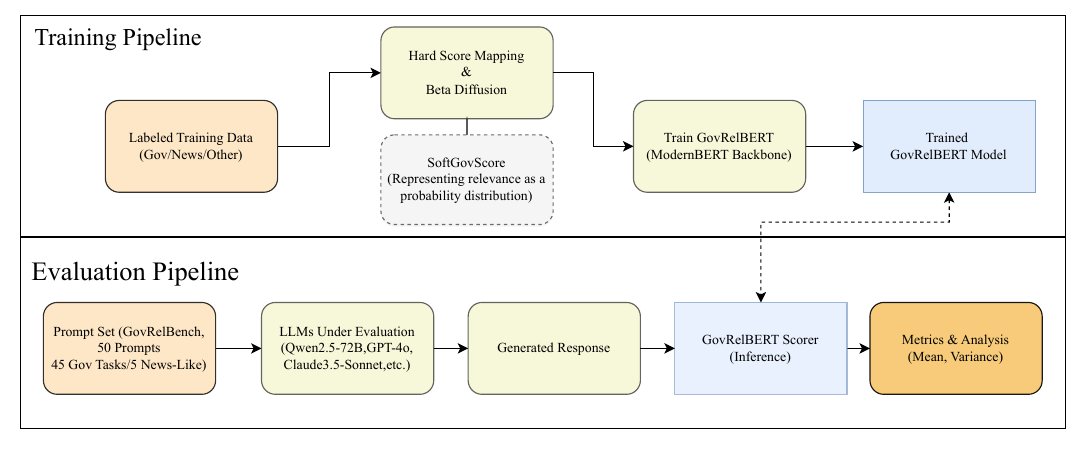}
    \caption{Overall pipeline of GovRelBench and GovRelBERT scoring workflow} \label{fig1}
\end{figure*}
We propose GovRelBench, a benchmark to evaluate the capabilities of LLMs in the government domain. This benchmark comprises the test prompts and a dedicated evaluation tool. To evaluate an LLM, prompts from GovRelBench are input to it, and our evaluation tool is then used to assess the relevance of the generated output to the government domain. This relevance assessment serves to measure the LLM's performance in this domain.

Initially, our evaluation tool employed traditional machine learning to perform binary classification on output text (determining if it pertained to the government domain). However, due to the overlap between texts from the government and news domains, this task requires more than simple classification or regression. It necessitates a specialized label processing approach, enabling the model to learn finer-grained relevance measures, ideally yielding soft labels~\cite{galstyan2007empirical} or continuous scores. Encoder-based language models, exemplified by BERT~\cite{devlin2019bert}, are well-suited for such understanding-intensive tasks. We selected ModernBERT~\cite{warner2024smarter} as the foundational architecture for our evaluation tool, primarily due to its sufficient context window capacity.For further details on the model architecture, please refer to the Appendix~\ref{Appendix 0}.

Our contributions are summarized as follows:

\begin{itemize}

    \item \textbf{GovRelBench}: A benchmark for evaluating the capabilities of LLMs in the government domain, comprising the prompts and an evaluation tool.

    \item \textbf{GovRelBERT}: A ModernBERT-based model for computing the government domain relevance score of given text, serving as the evaluation tool for GovRelBench.

    \item \textbf{SoftGovScore}: A method for representing text relevance that effectively converts "hard labels" to "soft scores," enabling more precise quantification of a text's relevance to the government domain, and is used for training GovRelBERT.
\end{itemize}

\section{Related Work}
\label{section:2}

\subsection{Soft Labels}
Soft labels represent a classification method in machine learning that offers finer granularity compared to hard labels~\cite{galstyan2007empirical}. They allow the assignment of probabilities or membership degrees to each category for a given data point. Hard labels, typically represented in a one-hot encoded format, explicitly assign a data point to a single category (represented by 1, with all others as 0). In contrast, soft labels can represent the likelihood that a data point belongs to multiple categories simultaneously, for instance, assigning a probability of 0.7 to category A and 0.3 to category B.

This approach proves particularly valuable when dealing with uncertainty, noisy data, and in semi-supervised learning scenarios. Studies indicate that soft labels can serve as a form of regularization, preventing the model from becoming overconfident and thereby enhancing its generalization capability on unseen data. Common application scenarios for soft labels include environments characterized by annotation uncertainty, high proportions of noisy data, challenges with model generalization, and knowledge distillation. For example, soft labels have been proposed to improve binary classification models in medical applications, particularly in situations where expert confidence is low~\cite{nguyen2014learning}. They also facilitate knowledge distillation, where the soft outputs (predictions) from a teacher model are used as soft labels to train a student model~\cite{hinton2015distilling}. 

Furthermore, various methods exist for obtaining soft labels, such as dynamically learning them via meta-learning~\cite{vyas2020learning} and collecting them through crowdsourcing~\cite{collins2022eliciting}.

\subsection{Open Source Datasets}
The thriving research ecosystem surrounding LLMs has led to the emergence of numerous open-source datasets. This study primarily focuses on datasets relevant to LLMs pre-training.

\textbf{MNBVC},contributed by the Liwu community~\cite{mnbvc}, is a comprehensive dataset that includes not only pre-training text data but also question-answering pairs and multi-turn dialogue data. It provides bilingual (Chinese-English) parallel corpora and features fine-grained categorization of data by domain.

\textbf{WuDaoCorpora},developed by the Beijing Academy of Artificial Intelligence (BAAI)~\cite{c6a3fe684227415a9db8e21bac4a15ab}, is a large-scale multimodal dataset. It encompasses text, dialogue, image-text pairs, and video-text pairs. The final dataset was curated by applying over 20 sophisticated cleaning rules to an initial corpus of 100TB of raw web data.

\textbf{IndustryCorpus2},also developed by BAAI~\cite{beijing_academy_of_artificial_intelligence}, is primarily designed for fine-tuning industry-specific models. It stands out by offering not only detailed categorization based on data sources but also annotations regarding data quality. Furthermore, it provides comprehensive statistical information about the text data, enabling researchers to gain a clearer understanding of its characteristics.

In addition to these examples, numerous other valuable pre-training datasets have been developed and made available to the research community.

\subsection{ModernBERT}
BERT, introduced by Google in 2018~\cite{devlin2019bert}, is a pre-trained language model based on the Transformer architecture and trained on large-scale text corpora. It achieves pre-training through two core tasks: Masked Language Modeling (MLM) and Next Sentence Prediction (NSP). While the original BERT revolutionized NLP development in 2018, its sequence length limit of 512 tokens constrained its ability to process long texts, and its computational efficiency suffered on longer sequences. Subsequent models like RoBERTa~\cite{liu2019roberta}, DeBERTa~\cite{he2020deberta}, and ALBERT~\cite{lan2019albert} attempted to improve performance through training on more data, employing decoupled attention mechanisms, or enhancing parameter efficiency, yet they still faced challenges regarding long sequence processing and computational demands.

ModernBERT, proposed in 2024~\cite{warner2024smarter}, addresses this gap, specifically designed for long context handling and efficient inference. Key features and advantages include:

\begin{itemize}
    \item \textbf{Extended Sequence Length}: ModernBERT supports a sequence length of 8192 tokens, a significant increase compared to BERT's 512 tokens, enabling better processing of longer documents.
    \item \textbf{Enhanced Activation Function}: It utilizes the GeGLU~\cite{shazeer2020glu} activation function, replacing traditional ReLU or GELU, to enhance the performance of its feed-forward layers.
    \item \textbf{Improved Efficiency with Unpadding}: It employs 'unpadding' to avoid computational waste on padding tokens, thereby boosting efficiency, especially in mixed-length batches.
    \item \textbf{Optimized Inference}: Its inference is optimized for common GPUs, making it suitable for real-world deployment scenarios, particularly those with limited resources.
\end{itemize}

ModernBERT demonstrates strong performance across various evaluations, including classification tasks and single/multi-vector retrieval, covering diverse domains such as code.

\section{Methology}
\subsection{SoftGovScore}
\label{section:3.1}
\textbf{Hard Labels to Hard Scores: Quantifying  Relevance}
Our initial efforts focused on binary classification to categorize text as either 'governmental' or 'non-governmental'. However, standard techniques like TF-IDF with Random Forest and embedding-based methods proved ineffective. The primary challenge stems from the fuzzy boundaries of the governmental domain, as its data frequently overlaps with other areas, making distinct separation difficult.

For instance, government activities reported online often fall into the news category (showing strong relevance), while documents from specific departments (e.g., education, finance) relate to both governance and their respective fields (showing weaker relevance). Recognizing that relevance is more of a spectrum than a strict binary choice, we abandoned direct domain classification. Instead, we adopted an approach focused on quantifying the degree of relevance using a numerical score, such as $Government = 1$, $News = 0.7$ and $Other = 0$, to better capture these nuances.

Let L be the initial domain label and H be the hard score. The conversion function $f_{\text{map}}$
  maps the domain label to a hard score(the mapping is detailed in Appendix~\ref{Appendix A}):
\begin{equation}
    H_L = f_{\text{map}}(L)
    \label{eq:hard_score}
\end{equation}
where $H \in [0, 1]$ for example, $H \in \{0, 0.7, 1, \dots\}$.

\textbf{Hard Scores to Soft Scores: Diffusion Score}
Building on the need for numerical relevance scores (Section \ref{section:3.1}, we convert initial domain labels into "hard scores" within the $[0, 1]$ range based on assessed relevance to the governmental domain. However, relying solely on these discrete hard scores (e.g., assigning exactly 1,  0.7, or 0) can force model predictions into these rigid points and inadequately represent the spectrum of relevance. Therefore, we diffuse these hard scores into probabilistic "soft scores".

\textbf{Selecting Distribution Function}
Our initial exploration involved using Gaussian distributions centered on each hard score. While this method can disperse scores and mitigate clustering effects, it introduces significant limitations. The application of Gaussian distributions necessitates the selection of a variance. However, assuming the same variance (i.e., level of uncertainty) for diffusing a high relevance score (e.g., 0.9) and a medium relevance score (e.g., 0.5) is problematic. Consequently, we opted for the \textbf{Beta distribution}. The Beta distribution allows for consistent prior confidence across different diffusion functions while maintaining manageable variance. Further analysis is detailed in Appendix ~\ref{Appendix D}.

\begin{equation}
    S_{\text{soft}} \sim \beta(\alpha, \beta)
    \label{eq:soft_score_dist}
\end{equation}

\begin{equation}
    P(S_{\text{soft}} \mid \alpha, \beta) = \frac{\Gamma(\alpha + \beta)}{\Gamma(\alpha)\Gamma(\beta)} S_{\text{soft}}^{\alpha-1} (1 - S_{\text{soft}})^{\beta-1}
    \label{eq:soft_score_pdf}
\end{equation}
Eq. (\ref{eq:soft_score_dist}) indicates that our soft scores follow a Beta distribution, while Eq. (\ref{eq:soft_score_pdf}) defines the probability density function (PDF) governing our score diffusion.

By utilizing the formula for the expectation of the Beta distribution, we determine the alpha and beta parameters for each class such that the following equation is satisfied:
\begin{equation}
    \mathbb{E}[beta_L(\alpha, \beta)]=\frac{\alpha}{\alpha+\beta}=H_L
\end{equation}

\subsection{GovRelBERT}
We constructed the training dataset for our evaluation model using self-crawled data combined with specific domain data filtered from open-source datasets, to which the SoftGovScore method was applied. Then we trained GovRelBERT, employing a pre-trained multilingual ModernBERT as the base model and using this dataset. Mean Squared Error (MSE \ref{eq:mse_loss}) was utilized as the loss function for the model's text relevance computation.
\begin{equation}
    L_{\text{MSE}} = \frac{1}{N} \sum_{i=1}^{N} (S_{\text{pred},i} - S_{\text{soft},i})^2
    \label{eq:mse_loss}
\end{equation}
\subsection{GovRelBench}

We constructed GovRelBench, a benchmark for evaluating the performance of LLMs in the government domain. This benchmark comprises 50 prompts and an evaluation tool (GovRelBERT). Among the 50 prompts, 45 are instructions or questions authored by us, designed to cover various aspects of the contemporary government domain. The remaining 5 prompts are derived from journalistic inquiries regarding government affairs, selected for their ability to be answered without requiring additional background information, thus ensuring broader applicability. The objective is to ensure coverage of typical scenarios within the government domain.
\begin{figure*}[t]
    \includegraphics[width=\textwidth]{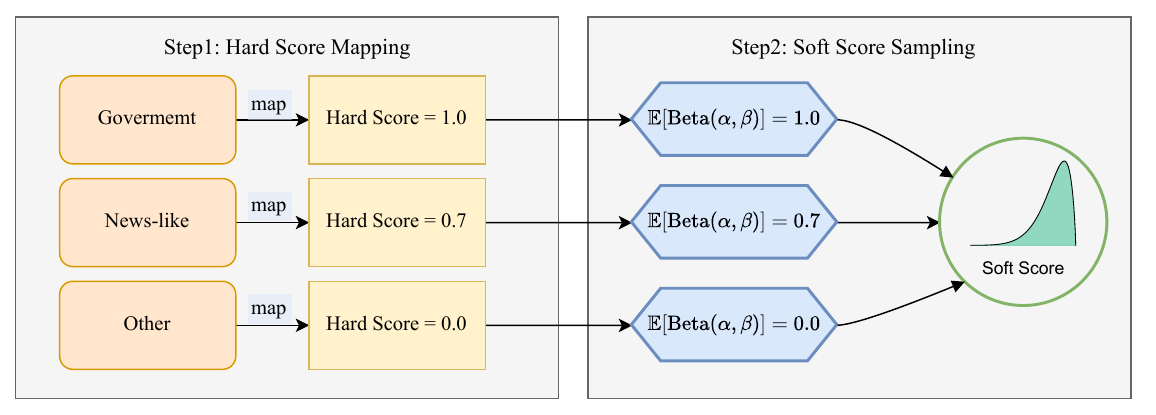}
    \caption{Two-step Construction of SoftGovScore: Category Mapping (Step 1) and Beta Diffusion (Step 2)} \label{fig2}
\end{figure*}

Although LLMs operate somewhat like "black boxes," their outputs, generated through inference and decoding processes, can indirectly reflect underlying tendencies, knowledge distribution, or lexical preferences concerning a specific domain, such as the government sector. GovRelBench leverages this: after an LLM generates text based on an input prompt, the GovRelBERT model is employed as an automated scoring tool to assess the relevance of the LLM-generated text to the government domain. This procedure provides a quantitative metric for evaluating an LLM's performance in the government domain.
\begin{table*}[ht]
    \centering
    \caption{Comparison of Evaluation Metrics for Different Models}
    \label{tab1}
    \begin{tabular}{lcc}
        \toprule
        \textbf{Model Name} & \textbf{Accuracy} & \textbf{F1 Score} \\
        \midrule
        TF-IDF + Random Forest & 34.44\% & 0.2750 \\
        M3e + LDA & 50.56\% & 0.3054 \\
        M3e + Logistic Regression & 49.4\% & 0.2980 \\
        Baseline (ModernBERT) & 24.36\% & 0.2625 \\
        Qwen2.5-7B-Instruct & 35.15\% & 0.3147 \\
        GLM-4-9b-chat & 30.92\% & 0.2753 \\
        InternLM2.5-7b-chat & 35.92\% & 0.3378 \\
        Hunyuan-A13B-Instruct & 40.41\% & 0.3617 \\
        Qwen2.5-14B-Instruct & 40.95\% & 0.4036 \\
        Qwen2.5-32B-Instruct & 41.83\% & 0.3739 \\
        \midrule
        $GovRelBERT_A$ (Error Tolerance 0.1) & \textbf{71.47}\% & \textbf{0.7394} \\
        $GovRelBERT_A$ (Error Tolerance 0.2) & \textbf{93.58}\% & \textbf{0.9380} \\
        $GovRelBERT_B$ (Error Tolerance 0.1) & \textbf{66.93}\% & \textbf{0.6929} \\
        $GovRelBERT_B$ (Error Tolerance 0.2) & \textbf{92.19}\% & \textbf{0.9250} \\

        \bottomrule
    \end{tabular}
\end{table*}

\section{Experiment}
\subsection{Experimental Setup}
\label{section:4.1}
The base model used for training GovRelBERT was 'modern\_bert\_multilingual' from 'neavo', a specific implementation of the ModernBERT architecture. Model training was conducted on an NVIDIA RTX 3090 GPU. The open-source datasets utilized for preparing our training data were sourced from the following Hugging Face repositories: 'BAAI/IndustryCorpus2', 'p208p2002/wudao', and 'liwu\slash MNBVC'. To ensure reproducibility, the random seed for all conducted experiments was set to 3407.
\subsection{Data Processing Implementation}
\label{section:4.2}
This section outlines the practical execution of our data processing methodology.

\textbf{Data Collection and Initial Sampling}
To account for potential overlaps between other domain data and the government domain, we collected data across 18 categories. Two categories, namely 'Government' and '$Law_A$', were self-collected. Unclassified data from 'wudao' was designated as the 'Other' category.
The remaining 15 domain categories were sampled from the large-scale open-source corpora referenced in Section \ref{section:4.1}: 'BAAI/IndustryCorpus2', 'p208p2002/wudao', and 'liwu/MNBVC'.

\textbf{Hard Score Assignment and Stratified Sampling} Following the quantified relevance proposed in Section \ref{section:3.1}, each category label was mapped to an initial "hard score" between 0 and 1 to reflect its estimated relevance to the government domain. This involved setting anchor points for clear cases(e.g., $Government~\approx 1$, $News~\approx 0.7$, $unrelated~\approx 0$) and subjectively assigning scores to intermediate categories, aiming to construct a smooth relevance spectrum. Subsequently, stratified sampling was performed based on these hard scores using score intervals of width 0.1  (e.g., $[0, 0.1)$, $[0.1, 0.2)$, etc.) to mitigate data imbalance issues. During an optimization step, boundary scores were slightly adjusted ('Government' to approximately 0.95 and 'Other' to approximately 0.15) to refine subsequent diffusion steps $H_L \to H_L^{\prime}$.

\begin{figure}[t]
    
    \includegraphics[width=0.9\columnwidth]{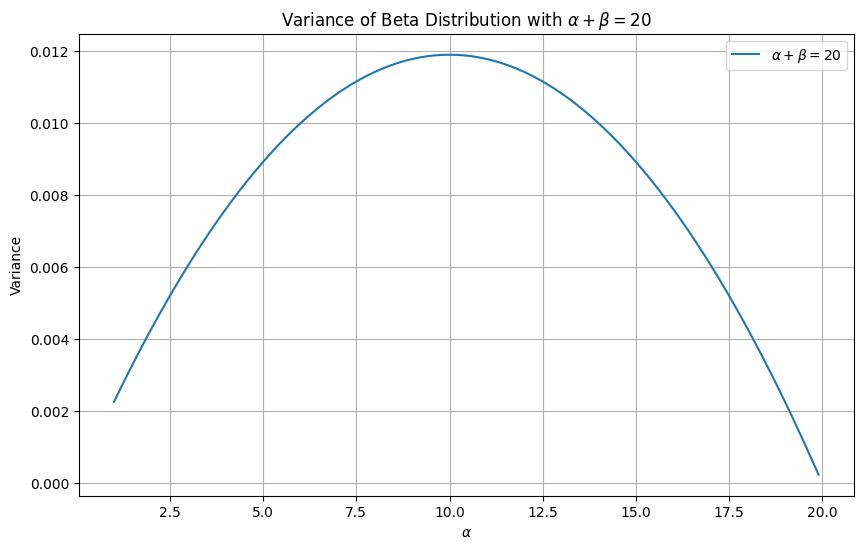}
    \caption{Variance Profile of the Beta Distribution $\alpha + \beta = 20$} \label{fig3}
    
\end{figure}

\begin{figure}
    
    \includegraphics[width=0.9\columnwidth]{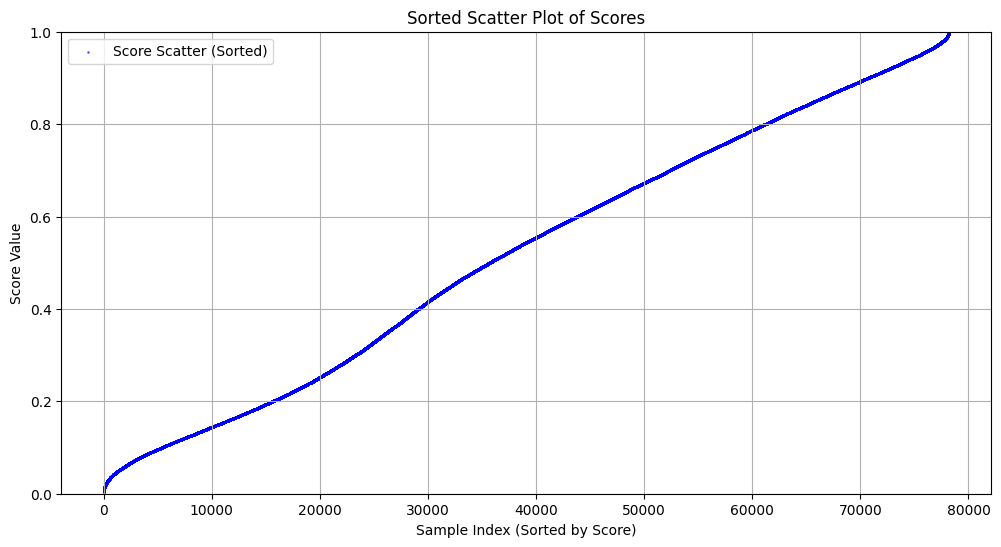}
    \caption{Reference Score Distribution} \label{fig4}
    
\end{figure}

\textbf{Soft Score Generation via Beta Diffusion}: Finally, we implemented the core "hard score to soft score" transformation using Beta distribution diffusion (Section \ref{section:3.1}). A consistency constraint of $\alpha+\beta = 20$ was applied to all categories. This ensured uniform prior confidence while allowing the score variance to adjust appropriately based on the hard score—resulting in lower variance near the boundaries (0/1) and higher variance in the mid-range (the variance distribution is shown in Fig. \ref{fig3}. The $\alpha$ and $\beta$ parameters are varied in increments of 0.5 for computational convenience. Therefore, in our practical implementation, we used $\frac{\alpha}{\alpha+\beta}= H_L^{\prime}$. This process generated the final soft score targets for model training, and their overall distribution is depicted in Fig. \ref{fig4}.

\subsection{GovRelBERT}
\label{section 4.3}
This section describes the training process and evaluation for our GovRelBERT model.
\subsubsection{Training Details}

The processed dataset comprises a total of 78,200 samples. Each sample in the dataset includes the following features: 'text' (the original source text), 'label' (the original domain category label from the source data), and 'score' (the target government domain relevance score generated via Beta diffusion, as described in Section~\ref{section:4.2}). This dataset was partitioned into training, evaluation, and test sets at a ratio of $8.5:1:0.5$, respectively. Key hyperparameters for training can be found in the code we provide. An early stopping strategy was employed based on Mean Squared Error (MSE) on the dataset. The optimal model, which we called $GovRelBERT_A$, was achieved at approximately 0.7 epochs, which may indicate data redundancy or high learning efficiency.

Furthermore, to validate the effectiveness of our proposed SoftGovScore method in directly regressing to soft scores, we conducted an ablation study. This involved training an alternative model,we called $GovRelBERT_B$, which employed an approach similar to traditional soft labeling: it first predicted probabilities for each original domain category and then converted these probabilities into relevance scores. Further details on this ablation study are provided in Appendix ~\ref{Appendix B}~.

\subsubsection{GovRelBERT Evaluation}

We assessed GovRelBERT ($GovRelBERT_A$) through three comparisons.

\textbf{Comparison with Baseline Models and Traditional Methods.} We benchmarked GovRelBERT against prior methods (TF-IDF+Random Forest, m3e+classifier~\cite{chen2024bge}) and a standard ModernBERT binary classifier (details in Appendix ~\ref{Appendix C}~). All these models were evaluated on a "government/non-government" binary classification task using the test set (approximately 3,500 samples) derived from the partitioning in Section \ref{section 4.3}. For GovRelBERT, which outputs continuous scores, its outputs were converted into categorical predictions by considering whether the absolute error from the target score was within thresholds of 0.1 and 0.2, respectively, to compute F1 score and accuracy metrics. As shown in \textbf{Table ~\ref{tab1}~}, GovRelBERT significantly outperformed the baseline models and traditional machine learning methods, demonstrating its capability in identifying government domain relevance.

\textbf{Comparison with pre-train LLMs.}
We also conducted a comparative analysis using several decoder-only Large Language Models (LLMs) for text classification. For this purpose, we selected a range of open-source models, including Qwen2.5, GLM-4, InternLM2.5, and Hunyuan, encompassing both dense and Mixture-of-Experts (MoE) architectures. Regarding model scale, we initially experimented with models of a comparable parameter count to ModernBERT, but found their instruction-following capabilities to be poor, rendering them almost incapable of producing coherent outputs. Consequently, we selected models ranging from 7B to 32B parameters for our evaluation. As shown in \textbf{Table ~\ref{tab1}~}, our model demonstrates superior performance compared to a 32B LLM, which has approximately 200 times more parameters. It is also noteworthy that the LLMs were tasked with classification, whereas our model performed a more nuanced regression-based scoring task.

\subsubsection{Comparison with Ablation Model}
\paragraph{Judgment of the panel of human.}
We compared $GovRelBERT_A$ (direct score regression) with $GovRelBERT_B$ (an ablation model employing a classify-then-score approach). Quantitatively, $GovRelBERT_A$ demonstrated superior performance within a tolerance of 0.1, whereas both models performed comparably within a tolerance of 0.2. In the qualitative assessment, a panel comprising one expert\footnote{A Senior Engineer with 11 years of experience in the governmental affairs.} and four graduate students reviewed samples exhibiting high and low errors. The panel indicated a 3:2 preference for $GovRelBERT_A$.

These findings suggest that: a) $GovRelBERT_A$'s direct regression approach more effectively captures fine-grained relevance, accounting for its superior performance within the 0.1 error tolerance; b) the comparable performance of both models at the 0.2 error tolerance reflects inherent noise or ambiguity in the data, a point substantiated by the analysis of high-error samples. This conclusion further underscores the effectiveness of our SoftGovScore method.

\paragraph{Judgment of the panel of LLMs.}
To address the limitation of relying on only five human annotators for the qualitative comparison, we adopted the recent \emph{LLM as a Judge} paradigm  and enlisted five state‑of‑the‑art open‑source large language models (LLMs) to act as additional independent evaluators.
Each LLM received the same paired outputs from $ModernBERT_A$ and $ModernBERT_B$  and produced a single‑sentence preference judgement.
Table~\ref{tab:llm-judge} summarises the aggregated results.

\begin{table}[t]
    \centering
    \caption{Pairwise preferences of five LLM judges (\checkmark = preferred, -- = tie).}
    \begin{adjustbox}{max width=\columnwidth}
    \begin{tabular}{lcc}
    \toprule
    \textbf{LLM Judge} & $ModernBERT_A$ & $ModernBERT_B$ \\
    \midrule
    Gemini\text{-}2.5\text{-}Pro & -- & -- \\
    Claude\text{-}Sonnet\text{-}4 &  & \checkmark \\
    GPT\text{-}4o & \checkmark &  \\
    Grok\text{-}4 & \checkmark &  \\
    DeepSeek\text{-}V3 &  & \checkmark \\
    \midrule
    \textbf{Majority Vote} & \textbf{2} & \textbf{2} \\
    \textbf{Voting ratio} & \textbf{11} & \textbf{9}\\
    \bottomrule
    \end{tabular}
    \end{adjustbox}
    \label{tab:llm-judge}
\end{table}

\textbf{Findings.}  
Two of the five LLM judges preferred $ModernBERT_A$, two preferred $ModernBERT_B$, and one returned a tie, resulting in an even 2–2 split by majority vote.  However, when counting individual preferences across all instances, $ModernBERT_A$ received 11 votes versus 9 for $ModernBERT_B$, indicating a slight overall lean towards the regression model. 

These automated judgements reinforce our human evaluation: although neither model dominates by a large margin, the marginal preference and moderate agreement both support the finer‑grained regression scoring of $ModernBERT_A$,.

\begin{table*}[ht]
    \centering
    \caption{Benchmark Scores for Different Models}
    \label{tab2}
    \begin{tabular}{lcc}
        \toprule
        \textbf{Model Name} & \textbf{Mean} & \textbf{Variance} \\
        \midrule
        deepseek-chat & 0.7959 & 0.0154 \\
        Qwen/Qwen2.5-72B-Instruct & \textbf{0.7967} & \textbf{0.0065} \\
        internlm/internlm2\_5-20b-chat & 0.77645 & 0.0099 \\
        TeleAI/TeleChat2-52B & 0.7459 & 0.0127 \\
        claude-3-7-sonnet-20250219 & 0.45328 & 0.0541 \\
        gpt-4o-2024-11-20 & 0.7359 & 0.0226 \\
        \bottomrule
    \end{tabular}
\end{table*}

\subsection{Benchmark Application and Evaluation}
To demonstrate the practical application of our GovRelBench benchmark and the GovRelBERT scoring model, we evaluated the performance of several prominent LLMs on governmental relevance. The models tested included DeepSeek-Chat, Qwen1.5-72B-Instruct, TeleChat-52B and so on.

The evaluation procedure was straightforward:
\begin{itemize}
    \item Each question/prompt from the GovRelBench dataset was directly inputted into the selected LLMs.
    
    \item The generated responses from each model were collected.
    
    \item Our trained GovRelBERT model ($GovRelBERT_A$) was then used to assess each response, assigning it a governmental relevance score based on the methodology described earlier.
\end{itemize}
This process allows for a quantitative comparison of how well different LLMs generate content relevant to the governmental domain when prompted with domain-specific questions.

As shown in Table \ref{tab2}, the benchmark results provide several insights into the performance of current general-purpose LLMs on Chinese governmental domain tasks. Among the models primarily developed in China, Qwen/Qwen1.5-72B-Instruct demonstrated the strongest performance, achieving the highest GovRelBench score in this group. When comparing prominent international models, GPT-4o exhibited a notably better understanding of Chinese governmental context, scoring significantly higher than Claude-3 Opus.

Furthermore, the score obtained by TeleAI/TeleChat-52B\footnote{TeleChat models are developed by China Telecom and are often referred to as public sector models, as they are pre-trained on the open-source TeleChat-PTD dataset, which is rich in governmental affairs data.} might reflect the impact of its parameter size. It is plausible that a larger version of this model, such as a hypothetical 115B parameter variant, could potentially achieve a higher relevance score on GovRelBench.

\section{Limitations}
The data preparation process, encompassing the collection of relevant texts and the initial mapping of domain labels to numerical relevance scores, involves procedures that lack universal applicability and thus relies heavily on subjective judgment and considerable domain expertise. Additionally, the inherent noise within the large-scale source datasets used for training poses an unavoidable challenge that may affect model performance. Our SoftGovScore method has currently only been validated in the government domain and has not been extended to more domains. In the future, we plan to apply our method to additional domains to verify its generalizability and will also attempt to develop more objective mechanisms for quantifying relevance to mitigate subjectivity.

\section{Conclusion}
In this paper, we introduced GovRelBench, a novel benchmark explicitly designed for evaluating the capabilities of large language models (LLMs) in the government domain. GovRelBench consists of carefully curated test prompts and an accompanying evaluation model, GovRelBERT, which quantifies the relevance of generated text to the government domain.

Our GovRelBERT model builds upon the ModernBERT architecture and leverages the proposed SoftGovScore method, a two-step label transformation technique that effectively converts categorical domain labels into nuanced soft relevance scores. Through comprehensive experiments, GovRelBERT demonstrated superior performance compared to traditional machine learning baselines, standard encoder-based classifiers, and even much larger decoder-only LLMs. Moreover, comparative analysis between direct regression and classify-then-score strategies highlighted the advantage of directly modeling continuous relevance scores.

Future research will aim to expand the generalizability of SoftGovScore to other specialized domains and address the inherent subjectivity and data quality challenges present in constructing domain-specific relevance measures. Additionally, we envision developing automated approaches for more objectively quantifying domain relevance, thereby enhancing the interpretability and reliability of benchmarks like GovRelBench.

\bibliography{custom}

\appendix

\section{Appendix}
\subsection{Explanation of chosen ModernBERT}
\label{Appendix 0}
For a benchmark's evaluation tool, the primary considerations are accuracy, reliability, speed, and deployment cost. We initially considered training a traditional machine learning classifier and explored various methods, including dimensionality reduction like embedding models. However, these approaches yielded low accuracy. Furthermore, our requirement is for a governmental affairs relevance score rather than a hard classification.

This led us to focus on models with strong natural language understanding capabilities. The mainstream paradigms are encoder-based models, represented by BERT, and decoder-based models, represented by GPT. Since decoder models typically have a larger parameter count and are better suited for natural language generation tasks, they are more appropriate for ranking the relative governmental relevance of several texts rather than assigning specific scores. Consequently, we selected an encoder-based model. The training task for an encoder model allows it to see the full context of the text and makes it more suitable for regression-style tasks.

Considering that texts in our scenario are often quite long, we ultimately chose ModernBERT as our backbone. We selected ModernBERT-base as our foundation model, which, with a parameter count of only 149M, effectively meets our requirements for inference speed and deployment cost.
\subsection{Dataset Details}
\label{Appendix A}
Our dataset encompasses a total of 18 distinct categories. The table below (Table ~\ref{tab3}~) provides details on the data distribution across these categories and outlines the initial mapping from each category label to its assigned reference governmental relevance score (the "hard score").

It is important to reiterate that these tabulated scores represent the initial reference points. The final soft scores used for training GovRelBERT were generated by applying Beta distribution diffusion centered around these reference scores, subsequent to minor smoothing adjustments described in Section \ref{section:4.2}.

\begin{table}[ht]
    \small
    \centering
    \caption{Distribution and Scores of Different Categories in the Dataset}
    \label{tab3}
    \begin{tabular}{lcc}
        \toprule
        \textbf{Name} & \textbf{Count} & \textbf{Score} \\
        \midrule
        Government Affairs & 10000 & 1.00 \\
        Law\_A & 3000 & 0.90 \\
        News & 4000 & 0.70 \\
        Corporate Annual Reports & 700 & 0.60 \\
        Diplomacy & 1000 & 0.85 \\
        Soy Milk & 3000 & 0.10 \\
        Government Work Reports & 1500 & 0.85 \\
        Agriculture & 4000 & 0.60 \\
        Politics & 5000 & 0.80 \\
        Electric Power & 2000 & 0.55 \\
        Entertainment & 9000 & 0.10 \\
        Economy & 3000 & 0.65 \\
        Law\_B & 2000 & 0.80 \\
        Real Estate & 4000 & 0.55 \\
        Education & 6000 & 0.60 \\
        Science & 2000 & 0.35 \\
        Study Strong Country & 5000 & 0.65 \\
        Others & 12000 & 0.00 \\
        \bottomrule
    \end{tabular}
\end{table}

\subsection{$GovRelBERT_B$ Details}
\label{Appendix B}
Following the training of$GovRelBERT_A$, we trained $GovRelBERT_B$ as an ablation study to compare our direct score regression method (SoftGovScore) against an alternative approach. Leveraging the fact that our training data retains the original domain category labels (18 distinct classes), $GovRelBERT_B$ was designed with a dual-objective training process.

Specifically, $GovRelBERT_B$ was trained to perform two tasks simultaneously using the final hidden state representation (dimension 768 from the ModernBERT encoder):

\begin{itemize}
\item Domain Classification: Predict the original category label (out of 18) for the input text.
\item Relevance Score Regression: Predict the target governmental relevance score (the soft score).
\end{itemize}

This contrasts with $GovRelBERT_A$, which had a single regression head mapping the hidden state directly to the score (768 -> 1). $GovRelBERT_B$ effectively employed two heads branching from the same encoder output: a classification head (outputting 18 logits) and a regression head (outputting 1 score). The underlying hypothesis was that forcing the model to also learn the discrete domain categories might potentially aid or influence the learning of the nuanced relevance score.

To train $GovRelBERT_B$, we used a combined loss function that equally weighted the losses from both tasks:

\begin{itemize}
\item The classification task used CrossEntropyLoss.
\item The score regression task continued to use MSELoss.
\end{itemize}

The final loss formula was:

\begin{align*}
    loss = \frac{1}{2} \Big( 
        &\text{CrossEntropy}(\text{hidden\_states}, \text{label}) \\
        &+ \text{MSELoss}(\text{predict\_score}, \text{score}) 
    \Big)
\end{align*}

All other training hyperparameters (learning rate, number of epochs, weight decay, evaluation steps, etc.) were kept identical to those used for training $GovRelBERT_A$ to ensure a fair comparison between the two approaches.

\subsection{modernbert baseline}
\label{Appendix C}
The calculation for the ModernBERT baseline referenced in the main text originated from an initial experimental design focused on binary classification ('Governmental' vs. 'Non-Governmental'). The approach involved:
\begin{itemize}
    \item Obtaining the text embedding from the ModernBERT model's encoder output.
    \item Defining or obtaining representative embeddings for the two target classes: 'Governmental' and 'Non-Governmental'.
    \item Calculating the cosine similarity between the input text's embedding and each of the two class embeddings.
    \item Applying a softmax function to these two similarity scores to convert them into probabilities.
    Assigning the text to the class with the higher probability.
\end{itemize}
This procedure essentially mimics the manual construction of a classification 'head' based on semantic similarity in the embedding space, rather than using a standard, end-to-end trained classification layer. Our results indicated that this baseline approach did not yield effective performance for the binary classification task.

\subsection{Advantages of the Beta Distribution Function}
\label{Appendix D}
To address these issues, we adopted the Beta distribution. Its properties align well with our needs:
\begin{itemize}
\item \textbf{Bounded Support}: The Beta distribution is naturally confined to the $[0, 1]$ range, matching our score requirements without needing adjustments.
\item \textbf{Principled Confidence and Variance Control}: By fixing the sum of its parameters $\alpha + \beta$, we maintain a consistent level of prior confidence across all initial hard scores used as the mean. Crucially, with $\alpha + \beta$ fixed, the distribution's variance automatically adapts based on the mean (the hard score).
\item \textbf{Intuitive Variance Behavior}: The variance is lowest near the boundaries (0 and 1) and highest near the middle (0.5). This means higher certainty for strongly relevant or irrelevant categories, and lower certainty for intermediately relevant ones – a desirable trait for modeling domain relevance.
\item \textbf{Flexibility}: This approach is adaptable for more complex scenarios involving finer-grained labels or cross-domain topics.
\end{itemize}

Thus, using Beta distribution diffusion allows us to transform discrete hard scores into nuanced soft scores that better capture the uncertainty inherent in assessing governmental domain relevance.
\subsection{Comparison of ModernBERT with LLMs}
The parameter size of ModernBERT itself is only 149M. We also selected an LLM of a comparable parameter scale, Qwen2.5-0.5B-Instruct (which also has an 8K context window), for comparison. The most significant difference lies in computational resource utilization: ModernBERT's training efficiency is approximately three times that of Qwen (0.5B), and its inference speed also surpasses Qwen's. Therefore, ModernBERT is better suited than this LLM for the task of relevance score computation.

\end{document}